\documentclass[letterpaper]{article} 
\usepackage{graphicx}
\usepackage{amsmath}
\usepackage{colortbl}
\usepackage{xcolor}
\usepackage{soul}
\sethlcolor{yellow}
\usepackage{svg}
\usepackage{tabularx}
\usepackage{pifont}
\usepackage{float} 
\usepackage{aaai25}  
\usepackage{times}  
\usepackage{helvet}  
\usepackage{courier}  
\usepackage[hyphens]{url}  
\usepackage{graphicx} 
\usepackage{multirow}
\usepackage{natbib}  
\usepackage{caption} 
\usepackage{multirow}
\usepackage{algorithm}
\usepackage{algorithmic}
\usepackage{verbatim}
\usepackage{pgfplots}
\usepackage{booktabs}
\usepackage{subcaption}
\pgfplotsset{compat=1.18} 
\usepackage{tikz}
\usetikzlibrary{matrix, positioning}
\nocopyright 
\usepackage{newfloat}
\usepackage{listings}
\usepackage{svg}

\DeclareCaptionStyle{ruled}{labelfont=normalfont,labelsep=colon,strut=off} 
\floatstyle{ruled}
\newfloat{listing}{tb}{lst}{}
\floatname{listing}{Listing}
\newcommand{\cmark}{\textcolor{green}{\ding{51}}} 
\newcommand{\xmark}{\textcolor{red}{\ding{55}}}   

\title{MedCodER: A Generative AI Assistant for Medical Coding}
\author {
   Krishanu Das Baksi\equalcontrib\textsuperscript{\rm 1},
   Elijah Soba\equalcontrib\textsuperscript{\rm 2},
   John J. Higgins\textsuperscript{\rm 2},
   Ravi Saini\textsuperscript{\rm 1},
   Jaden Wood\textsuperscript{\rm 2},
   Jane Cook\textsuperscript{\rm 2},
   Jack Scott\textsuperscript{\rm 2},
   Nirmala Pudota\textsuperscript{\rm 1},
   Tim Weninger\textsuperscript{\rm 3},
   Edward Bowen\textsuperscript{\rm 2},
   Sanmitra Bhattacharya\textsuperscript{\rm 2}
}
\affiliations {
   \textsuperscript{\rm 1}Deloitte \& Touche Assurance \& Enterprise Risk Services India Private Limited\\
   \textsuperscript{\rm 2}Deloitte \& Touche LLP\\
   \textsuperscript{\rm 3}University of Notre Dame\\
}

\begin{document}

\maketitle

\begin{abstract}
Medical coding is essential for standardizing clinical data and communication but is often time-consuming and prone to errors. Traditional Natural Language Processing (NLP) methods struggle with automating coding due to the large label space, lengthy text inputs, and the absence of supporting evidence annotations that justify code selection. Recent advancements in Generative Artificial Intelligence (AI) offer promising solutions to these challenges. In this work, we introduce MedCodER, a Generative AI framework for automatic medical coding that leverages extraction, retrieval, and re-ranking techniques as core components. MedCodER achieves a micro-F1 score of 0.60 on International Classification of Diseases (ICD) code prediction, significantly outperforming state-of-the-art methods. Additionally, we present a new dataset containing medical records annotated with disease diagnoses, ICD codes, and supporting evidence texts (\url{https://doi.org/10.5281/zenodo.13308316}). Ablation tests confirm that MedCodER's performance depends on the integration of each of its aforementioned components, as performance declines when these components are evaluated in isolation.

\end{abstract}

%

\section{Introduction}

The International Classification of Diseases (ICD)\footnote{\url{https://www.cms.gov/medicare/coding-billing/icd-10-codes}}, developed by the World Health Organization (WHO)\footnote{\url{https://www.who.int/standards/classifications/classification-of-diseases}}, is a globally recognized standard for recording, reporting, and monitoring diseases. In the United States, the use of ICD codes is mandated by the U.S. Department of Health and Human Services (HHS) for entities covered by the Health Insurance Portability and Accountability Act for insurance purposes.

ICD codes have undergone various revisions over time to reflect advancements in medical science\footnote{\url{https://www.cdc.gov/nchs/hus/sources-definitions/icd.htm}}. The 10th revision, known as ICD-10-CM (referred to as ICD-10 hereafter) in the U.S, is the standard for modern clinical coding and comprises over 70,000 distinct codes. These codes follow a specific alphanumeric structure \cite{hirsch2016icd} and are organized into a hierarchical ontology based on the medical concepts they represent. For example, the code ``I50.1'' corresponds to ``Left ventricular failure, unspecified,'' where I50 represents "Heart Failure," and the ".1" indicates its specificity. ICD-10 differs significantly from previous versions, making translation between versions challenging.

\begin{figure*}[t]
\centering
\includegraphics[width=1.0\textwidth]{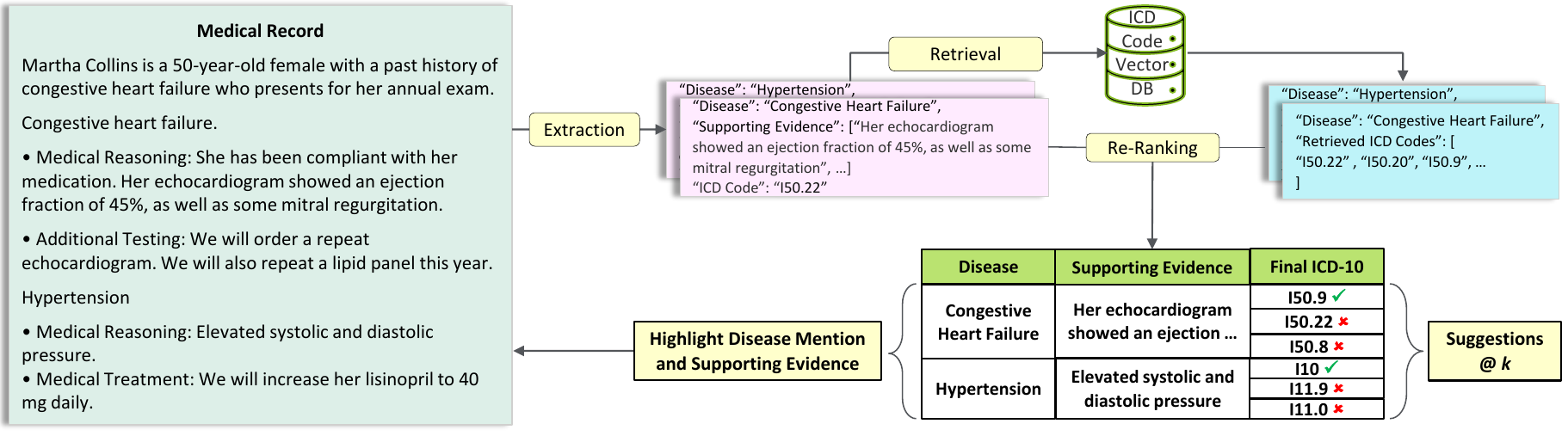}
\caption{A schematic diagram of the MedCodER framework illustrates three primary components: extraction of disease diagnoses,  supporting evidence and an initial list of ICD-10 codes, retrieval of candidate ICD-10 codes for the extracted diagnosis using a vector database, and re-ranking these combined codes to produce a final list of $k$ ICD-10 codes. Extracted disease mentions and supporting evidence are mapped back to the medical record for in-context highlighting, aiding medical coders in the coding process.}
\label{fig:method}
\end{figure*}

Accurate ICD coding is essential for medical billing, health resource allocation, and medical research \cite{campbell2020computer}. This task is performed by specialized professionals known as medical or clinical coders, who use a combination of manual techniques and semi-automated tools to process large volumes of medical records. Their primary responsibility is to accurately assign ICD-10 codes to medical records based on documented diagnoses and procedures. The coding process is often time-consuming and costly, with its difficulty varying depending on the complexity of the cases and the level of detail in the documentation. Errors in ICD coding can have significant financial and legal implications for patients, healthcare providers, and insurers. Despite the critical importance of accurate coding, few reliable solutions exist to supplement or automate this process.

Automation of ICD coding is an active research area within the NLP community. While various approaches have been proposed, recent methods typically frame this task as a multi-label classification problem: given a raw medical record text, the goal is to predict each of the relevant ICD codes \cite{yan2022survey}. Although the objective is straightforward, several challenges make automatic ICD coding difficult. These include the extremely large label space, the diversity and lack of standardization in medical record data, and the severely imbalanced distribution of labels. Advanced deep learning techniques have significantly improved the performance of automated ICD coding, but they still fall short of fully automating the process. Furthermore, these methods often lack interpretability, making it difficult to provide justifications for the selected ICD codes.

In the realm of Generative AI, Large Language Models (LLMs) have shown remarkable capabilities in text generation and reasoning, particularly in zero-shot scenarios. However, early efforts to apply LLMs for automatic ICD coding have produced unsatisfactory results \cite{boyle2023automatedclinicalcodingusing, poor_coders}. We hypothesize that augmenting the intrinsic (parametric) knowledge of LLMs with complementary techniques, such as retrieval \cite{lewis2020retrieval} and re-ranking \cite{sun-etal-2023-chatgpt}, can significantly improve their accuracy in this domain.

Evaluation, benchmarking, and reproducibility of results for automatic ICD coding tools, particularly those based on Generative AI, are challenged by licensing terms that restrict processing by external models. For example, the widely-used Medical Information Mart for Intensive Care (MIMIC) datasets \cite{Johnson2023MIMIC} include medical records and associated ICD codes but prohibit data sharing with third parties such as Anthropic, Mistral AI and OpenAI \cite{boyle2023automatedclinicalcodingusing}. Additionally, these datasets lack annotations for supporting evidence, which are crucial for justifying ICD code selections.

To address the challenges associated with applying Generative AI approaches to ICD coding and the lack of third-party-friendly ICD coding datasets, this paper makes the following contributions:

\begin{enumerate}
    \item We introduce an open-source dataset designed for evaluating ICD coding methodologies, including those based on Generative AI. This dataset includes not only ICD-10 codes but also extracted diagnoses and supporting evidence texts, which facilitate the development and assessment of interpretable ICD coding methods.
    \item We describe the \textbf{Med}ical \textbf{Cod}ing using \textbf{E}xtraction, \textbf{R}etrieval, and re-ranking (\textbf{MedCodER}) framework, an accurate and interpretable approach to ICD coding that leverages LLMs along with retrieval and re-ranking techniques. MedCodER first extracts disease diagnoses, supporting evidence, and an initial list of ICD-10 codes from medical records. It then retrieves candidate ICD-10 codes using semantic search and re-ranks the combined codes from previous steps to produce the final ICD-10 code predictions.
    \item We evaluate the performance of the MedCodER framework compared to state-of-the-art (SOTA) methods using our dataset.
    \item We illustrate how MedCodER can be integrated as a Generative AI assistant for medical coders through a preliminary user interface (UI) design.
\end{enumerate}

\section{Related Research}
\label{related_work}

\subsection{Automatic ICD Coding}

Automated ICD coding is a challenging NLP problem, approached through rule-based \cite{kang2013using, farkas2008automatic}, traditional machine learning \cite{scheurwegs2016data, 
scheurwegs2017selecting}, and deep learning methods \cite{Ji_2024_unified}. Recent methods often treat it as a multi-label classification task, utilizing architectures like convolutional \cite{mullenbach-etal-2018-explainable, cao-etal-2020-clinical}, recurrent \cite{yu2019automatic, guo2020disease}, graph neural networks \cite{Wang_Xu_Gan_Li_Wang_Chen_Yang_Wang_Carin_2020}, and transformers \cite{huang-etal-2022-plm}. Although generative AI and LLMs have been explored for ICD coding \cite{boyle2023automatedclinicalcodingusing, poor_coders}, results have been mixed.

An analysis by Edin et al. \citeyear{Edin_2023} compared SOTA ICD coding models on MIMIC datasets and found that PLM-ICD \cite{huang-etal-2022-plm} excelled on MIMIC IV, but common ICD coding challenges persisted, with \textit{more than half} of ICD-10 codes misclassified. This suggests the potential of zero-shot models like LLMs for more reliable solutions.

LLM-based ICD coding research has yielded mixed outcomes. One study achieved only a 34\% match rate using a dataset from Mount Sinai \cite{poor_coders}, while an LLM-guided tree search method achieved competitive results \cite{boyle2023automatedclinicalcodingusing}, though it lacked transparency in code selection and was resource-intensive.

\subsection{Disease Extraction}
Disease extraction, a key component of both traditional medical coding and the MedCodER framework, involves identifying disease entities from medical records and is a form of Named Entity Recognition (NER) in biomedical NLP \cite{Durango2023Named}. While often overlooked in ICD coding methods, disease NER is crucial for accurate retrieval and re-ranking of ICD codes.

Domain-specific models like Bidirectional Encoder Representations from Transformers for Biomedical Text Mining (BioBERT) \cite{Lee2019BioBERT}, pre-trained on biomedical literature, achieve high F1 scores (86-89\%) on benchmark datasets but are more effective with data similar to their training sets. Recent advancements, such as Universal Named Entity Recognition (UniNER), which uses knowledge distillation from LLMs, and Generalist Model for Named Entity Recognition (GLiNER), a transformer-based model with fewer than 1 billion parameters, have shown competitive zero-shot performance on the National Center for Biotechnology Information (NCBI) Disease corpus \cite{zhou2024universalnertargeteddistillationlarge, zaratiana2023gliner}.

Unlike general NER, which may identify a broad range of disease mentions, ICD-10 extraction focuses on diagnosing diseases relevant for coding, reducing noise and minimizing errors in billing and documentation. Our approach targets precise disease extraction aligned with ICD-10 codes.

\subsection{Retrieval and Re-ranking}
While traditional NLP methods often frame automatic ICD coding as a multi-label classification task, it can also be approached as a retrieval and re-ranking problem. In this perspective, the goal is to retrieve the most relevant ICD codes for a given medical record and then re-rank them into a prioritized list. This approach addresses the challenge of dealing with large label spaces by filtering out irrelevant codes, resulting in a more manageable set of candidates. The retrieval and re-ranking strategy is similar to the Retrieval-Augmented Generation (RAG) paradigm used in generative AI workflows, which further supports the use of LLMs in automatic ICD coding.

Prior work has explored the retrieval and re-ranking paradigm using pre-trained ICD coding models~\cite{tsai2021modelingdiagnosticlabelcorrelation}. In this approach, the top $k$ most probable codes are selected from the pre-trained model and re-ranked based on label correlation. However, its effectiveness is limited by the retriever's ability to produce relevant codes within the top $k$. Embedding models have also been utilized to retrieve relevant codes for a given medical record~\cite{Niu2023Retrieve}. While promising, this approach is limited by the challenges of long input texts and lacks a clear rationale for ICD-10 code selections. In contrast, the MedCodER framework addresses these limitations by extracting disease-related text segments to enhance the retrieval of relevant ICD-10 codes.

\section{MedCodER Framework}
Here we introduce the MedCodER framework, which is illustrated in Fig.~\ref{fig:method}. MedCodER is an interpretable and explainable ICD coding framework comprised three components: (1) extraction, (2) retrieval, and (3) re-ranking. In this section, we describe each component and its relevance to ICD-10 coding.

\subsection{Step 1: Disease Diagnoses, Supporting Evidence \& ICD-10 Code Extraction}
The first step of MedCodER uses a LLM, specifically GPT-4\footnote{GPT-4 was selected as it outperformed other commercial and open LLMs on the US medical licensing exams dataset (MedQA) in the HELM leaderboard \url{https://crfm.stanford.edu/helm/lite/v1.2.0/#/leaderboard}.}, to extract disease diagnoses, supporting evidence text, and associated ICD-10 codes. Disease diagnoses refer to the clinical terms describing a patient's ailment, while supporting evidence includes additional information about the disease, such as test results and medications. We asked the LLM to generate structured JSON output of these entities from the medical record.

Drawing inspiration from Chain-of-Thought (CoT) prompting \cite{wei2022chain}, we asked the LLM to first reason about relevant text from the medical record before generating ICD-10 codes, mimicking the workflow of medical coders. The extracted diagnoses are used in the retrieval step, while the supporting text and generated ICD-10 codes are used in the re-ranking step. 

To address potential hallucinations, we implemented additional verification steps. For diagnoses, we used fuzzy matching to replace extracted terms with their exact or closest match from the medical record. For supporting evidence, which tend to be longer phrases compared to diagnoses, we substituted them with sentences from the medical record that have the highest BM25 similarity score. Error mitigation for the generated ICD-10 codes occurs in the re-ranking step (Step 3). 

\subsection{Step 2: ICD-10 Retrieval Augmentation}
Instead of relying solely on traditional deep learning models or the parametric knowledge of LLMs for ICD coding, we augmented the LLM's outputs from the previous step by generating a candidate set of ICD-10 codes through semantic search between extracted diagnoses and the descriptions of valid ICD-10 codes. This approach mitigates the large label space issue by reducing the number of potential codes to a more manageable set.

For the semantic search, we compiled textual descriptions of valid codes from the ICD-10 ontology and equivalent descriptions from the Unified Medical Language System (UMLS) Metathesaurus\footnote{https://www.nlm.nih.gov/research/umls/index.html}, providing accurate handling of medical synonyms. We then embedded these descriptions and tagged each code with metadata related to the ontology, such as chapter, block, and category~\cite{boyle2023automatedclinicalcodingusing}. During inference, disease diagnoses are embedded, and the top $k$ most similar ICD-10 codes based on cosine distance are retrieved for each diagnosis. This results in a ranked list of ICD-10 codes directly mapped to specific diagnoses, enhancing interpretability.

\subsection{Step 3: Code-to-Record Re-ranking}
In the final step, the retrieved codes from the Step 2 and those generated by the LLM are re-ranked to produce the final list of predicted ICD-10 codes. This re-ranking is performed using an LLM, but only the extracted diagnoses and supporting evidence are considered, allowing the LLM to prioritize based on relevant information. Additionally, the retrieved ICD-10 codes are compared against the valid (\textit{i.e.}, billable) list to avoid inclusion of hallucinated outputs. This re-ranking methodology and prompts are based on the RankGPT framework \cite{sun-etal-2023-chatgpt}, with modifications specific to ICD-10 coding.

\section{Experimental Methodology}

\subsection{Dataset}

Because current ICD coding benchmark datasets, like MIMIC III and IV, have restrictions on use with off-the-shelf, externally-hosted LLMs, and because they lack annotations of supporting evidence text, they cannot be used in typical Generative AI solutions. To address these challenges, we created a new dataset that extends the Ambient Clinical Intelligence Benchmark (ACI-BENCH) dataset \cite{aci-bench}. ACI-BENCH is a synthetic dataset containing 207 transcribed conversations that simulate doctor-patient interactions. Clinical notes were generated using an automated note-generation system that summarized dialogues transcribed using human transcription and automatic speech recognition tools. These notes were then reviewed and revised, as necessary, by medical domain experts to ensure their accuracy and realism, closely mimicking real-world clinical notes. The ACI-BENCH dataset does not incorporate linkages to structured data, including vital signs, order codes, and diagnosis codes.

We extended the ACI-BENCH dataset by manually annotating each clinical note with ICD-10 codes, disease diagnoses, and supporting evidence texts, with the help of an expert medical coder. Of the 207 clinical notes, three were deemed unworthy of coding. The remaining notes were coded in two batches: the first batch included 184 notes, 360 ICD-10 codes with diagnoses, and 737 supporting evidence texts, and is used to evaluate the results of various MedCodER components. The second batch, consisting of 20 notes, is intended for use in A/B testing the coding application with and without the AI assistant. 

\subsection{Methodology}
We evaluate the efficacy of MedCodER's components, focusing on disease diagnoses extraction and ICD-10 coding, using the extended ACI-BENCH dataset and comparing them with SOTA approaches. For disease NER, we compare our approach against BioBERT, SciSpacy \cite{neumann2019scispacy}, UniNER and GLiNER.  For ICD-10 code generation, we compare MedCodER against PLM-ICD, and Clinical Coder prompt and LLM Tree-Search~\cite{boyle2023automatedclinicalcodingusing}. Because most automatic ICD coding baselines produce a single ICD-10 code per diagnosis, we compare our $k$@1 results against these. We also demonstrate performance trade-offs with increasing values of $k$. For non-LLM baselines, we use publicly available pre-trained weights, and for LLM-based experiments, we use GPT-4. Supporting evidence extraction is a unique and challenging problem with no existing baselines. We report results from various prompting approaches and identify this as an area for future research.

\subsection{Metrics}
We report results with micro precision and micro recall for each sub-task. Consistent with current evaluation approaches for NER and ICD coding, we focus on micro metrics because, in extremely large label spaces, it is crucial to treat each instance equally rather than each class. This approach emphasizes the performance of our framework per document rather than per ICD-10 code.

To evaluate disease diagnoses extraction, we use set-based, exact-match metrics.
Our metric choice is motivated by the retrieval subtask. Because vector search is location-independent, we disregard text positions when computing extraction performance. Additionally, we treat exact matches case insensitively, differing from traditional NER evaluations.

\section{Results}
In this section, we present the results of both the baselines and our proposed framework.

\begin{table}[t]
\centering
\begin{tabular}{l | lll}
\toprule
\textbf{Model}     & \textbf{Recall}        & \textbf{Precision}         & \textbf{F1}          \\ \midrule
BioBERT                    & 0.44          & 0.07          & 0.12             \\
SciSpacy                   & 0.64          & 0.10          & 0.17             \\
UniNER                     & 0.67          & 0.11          & 0.19          \\
GLiNER                     & 0.78          & 0.15          & 0.25         \\
MedCodER & \textbf{0.85} & \textbf{0.81} & \textbf{0.83}    \\ \bottomrule
\end{tabular}
\caption{Disease diagnoses extraction results. Best results are highlighted in bold.}
\label{tab:with_ps}
\end{table}

\subsection{Disease Diagnoses and Supporting Evidence Extraction}
The results of disease diagnoses extraction are shown in Table~\ref{tab:with_ps}. 
We find that MedCodER's disease diagnoses extraction for ICD-10 coding outperforms other disease NER systems, validating our hypothesis that prompting for specific ICD-10 diagnoses is better for this task. Because disease extraction is central to the ICD coding framework, these results set an upper bound on ICD coding performance. Notably, generalist models such as GLiNER outperformed domain-specific models, likely due to differences between general disease NER and ICD-10 diagnosis extraction objectives.

\begin{table}[t]
\label{table:combined_res}
\centering
\begin{tabular}{l|lll}
\toprule
\textbf{Model}                       & \textbf{Recall} & \textbf{Precision} & \textbf{F1} \\ \midrule
PLM-ICD         & 0.57            & 0.31               & 0.40              \\
Clinical Coder Prompt & 0.52            & 0.32               & 0.40              \\
LLM Tree-Search        & 0.53            & 0.10               & 0.17              \\
MedCodER @1    & \textbf{0.63}   & \textbf{0.58}      & \textbf{0.60}     \\ \bottomrule
\end{tabular}
\caption{ICD-10 coding results for MedCodER compared to SOTA baselines. Best results are highlighted in bold.}
\label{tab:baseline_results}
\end{table}

Because this dataset is the first to include supporting evidence for ICD-10 codes and their associated diagnoses, we lacked a baseline for comparison. In our experiments with various prompting approaches, partial match recall ranged from 0.75 to 0.82, and precision ranged from 0.24 to 0.30 (detailed results are omitted due to space constraints). The low precision indicates that the model extracts some non-relevant evidence, potentially introducing errors in the re-ranking process where supporting evidence texts are used. Despite the low precision, our full framework results, presented below, suggest that the extracted supporting evidence aids re-ranking. This task is more nuanced and challenging than disease extraction, highlighting the need for performance improvements in future work.

\subsection{ICD-10 Coding}
Table~\ref{tab:baseline_results} presents MedCodER results when filtering for only the top ranked ICD-10 code per diagnosis. For baselines, we used the pre-trained weights of PLM-ICD on MIMIC IV from Edin et al. \shortcite{Edin_2023} and a 50-call limit for the LLM Tree-Search. These methods represent the SOTA deep learning \cite{Edin_2023} and generative AI based solutions \cite{boyle2023automatedclinicalcodingusing} for automatic ICD-10 coding. MedCodER outperforms these baselines, significantly enhancing ICD-10 coding performance while remaining interpretable. 

\subsection{Ablation Results}

\begin{figure}[t]
    \centering
    \includegraphics[width=\columnwidth]{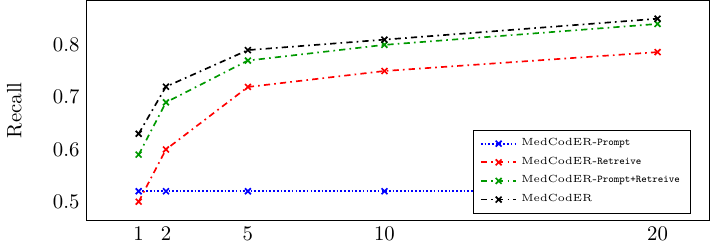}
    \includegraphics[width=\columnwidth]{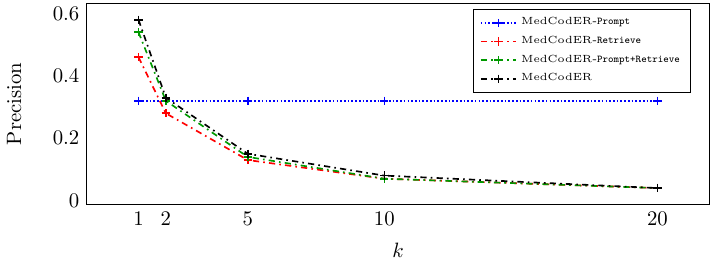}
    \caption{Recall and Precision @$k$ for  variations of MedCodER framework}
    \label{fig:ablation_figure}
\end{figure}

\begin{table}[t]
    \centering
    \begin{tabular}{>{\raggedright\arraybackslash}p{4.8cm} | p{3.0cm}}
        \toprule
        \textbf{Medical Record Snippet and Ground Truth Diagnosis } & \textbf{Ground Truth ICD-10 and Description}  \\
        \midrule
        \raggedright Regarding her \hl{depression}, the patient feels that it is well managed on Effexor & \textcolor{green}{F32.A}: Depression, unspecified \\
        \multicolumn{2}{c}{
        \begin{tabular}{@{}c@{}}
        \hfill
        \begin{subtable}[b]{.59\linewidth}                    
            \begin{tabular}{@{}|p{3.5cm}|p{1.5cm}|l|@{}}
                \raggedleft \textbf{Model} & \textbf{Prediction} & \textbf{?} \\ \hline
                \raggedleft MedCodER-\texttt{Prompt} & F32.9 & \xmark \\ 
                \raggedleft MedCodER-\texttt{Retrieve} & F33.9 & \xmark \\
                \raggedleft MedCodER & F32.A & \cmark \\
            \end{tabular}
        \end{subtable} 
        \end{tabular}} \\[1.0cm]        
        \midrule
        \raggedright Edema and ecchymosis surrounding the knee. Positive pain to palpation. Assessment: \hl{Right Knee Contusion} & \textcolor{green}{S80.01XA}: Contusion of right knee, initial encounter \\
        \multicolumn{2}{c}{
        \begin{tabular}{@{}c@{}}
        \hfill
        \begin{subtable}[b]{.59\linewidth}                    
            \begin{tabular}{@{}|p{3.5cm}|p{1.5cm}|l|@{}}
                \raggedleft \textbf{Model} & \textbf{Prediction} & \textbf{?} \\ \hline
                \raggedleft MedCodER-\texttt{Prompt} & S80.01XA & \cmark \\
                \raggedleft MedCodER-\texttt{Retrieve} & S80.01 & \xmark \\
                \raggedleft MedCodER & S80.01XA & \cmark \\
            \end{tabular}
        \end{subtable} 
        \end{tabular}} \\[1.0cm]   
        \midrule
        \raggedright Today I discussed conservative options for \hl{left shoulder impingement} with the patient & \textcolor{green}{M75.42}: Impingement syndrome of left shoulder \\ 
        \multicolumn{2}{c}{
        \begin{tabular}{@{}c@{}}
        \hfill
        \begin{subtable}[b]{.59\linewidth}                    
            \begin{tabular}{@{}|p{3.5cm}|p{1.5cm}|l|@{}}
                \raggedleft \textbf{Model} & \textbf{Prediction} & \textbf{?} \\ \hline
                \raggedleft MedCodER-\texttt{Prompt} & M75.40 & \xmark \\
                \raggedleft MedCodER-\texttt{Retrieve} & M75.42 & \cmark \\
                \raggedleft MedCodER & M75.42 & \cmark \\ 
            \end{tabular}
        \end{subtable} 
        \end{tabular}} \\[1.0cm] 
        \midrule
        \raggedright His examination is consistent with rather severe post-traumatic \hl{stenosing tenosynovitis of the right index finger}. &  \textcolor{green}{M65.321}: Trigger finger, right index finger \\ 
        \multicolumn{2}{c}{
        \begin{tabular}{@{}c@{}}
        \hfill
        \begin{subtable}[b]{.59\linewidth}                    
            \begin{tabular}{@{}|p{3.5cm}|p{1.5cm}|l|@{}}
                \raggedleft \textbf{Model} & \textbf{Prediction} & \textbf{?} \\ \hline
                \raggedleft MedCodER-\texttt{Prompt} & M22.40 & \xmark \\
                \raggedleft MedCodER-\texttt{Retrieve} & M17.2 & \xmark \\
                \raggedleft MedCodER & M22.2X1 & \xmark \\
            \end{tabular}
        \end{subtable} 
        \end{tabular}} \\
        \bottomrule
    \end{tabular}
    \caption{Error analysis of each variation of the MedCodER framework with associated disease diagnosis}
    \label{fig:errors}
\end{table}

\begin{figure*}[t]
    \centering
    \includegraphics[width=1.0\textwidth]{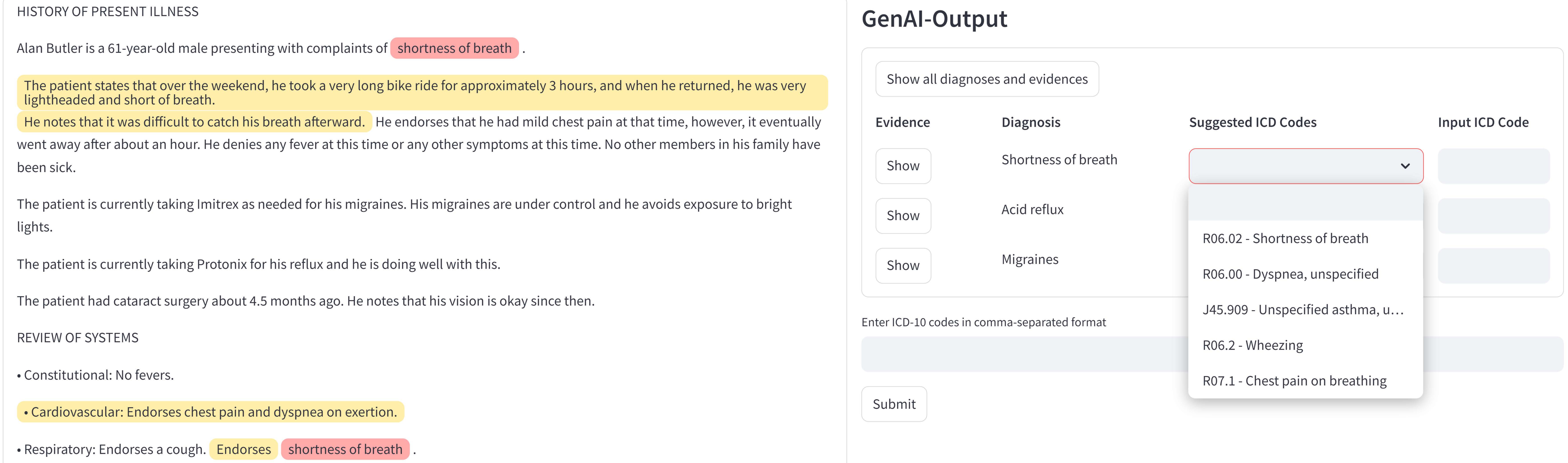}
    \caption{A representation of MedCodER in action. On the left, the medical record is annotated with the disease diagnosis for shortness of breath and its supporting evidence texts. On the right, the corresponding top 5 ICD-10 code suggestions are shown. Other diagnoses and supporting evidence texts can be toggled to show or hide using the `Show' buttons next to them.}
    \label{fig:main}
\end{figure*}

To evaluate the efficacy of retrieval and re-ranking on ICD coding performance, we conducted an ablation study. The results are shown in Fig.~\ref{fig:ablation_figure}. The variations of MedCodER used in the study are:
\begin{itemize}
    \item MedCodER-\texttt{Prompt}: Uses only the ICD-10 codes from prompting. Note that a single ICD code is generated per diagnosis by prompting and hence the performance of this variant does not change with value of $k$.
    \item MedCodER-\texttt{Retrieve}: Uses only the retrieved ICD-10 codes, without re-ranking.
    \item MedCodER-\texttt{Prompt+Retrieve}: Uses both prompted and retrieved ICD-10 codes, removing any non-valid or non-billable codes, without re-ranking.
    \item MedCodER: The entire framework with each constituent component, \textit{i.e.}, prompted and retrieved ICD-10 codes after re-ranking.
\end{itemize}

We observe that re-ranking the combined set of prompted and retrieved ICD-10 codes outperforms using either method alone. Recall increases monotonically with addition of retrieval, meaning our search produces semantically relevant hits. As expected, the precision decays as we produce more output codes. Contrary to prior work~\cite{poor_coders}, our results with MedCodER-\texttt{Prompt} show that LLMs can perform well on ICD-10 prediction with careful prompt engineering. We attribute this to prompt design, where the LLM is prompted to first generate the diagnoses and supporting evidence texts before it is prompted to generate the ICD-10 codes, akin to CoT prompting \cite{wei2022chain}. 

\subsection{Error Analysis}
We conducted an error analysis to highlight MedCodER's limitations and suggest future research directions.

Table \ref{fig:errors} presents failure cases for each component of our framework ($k$=1). We only show cases where the extracted disease diagnosis matched the ground truth to highlight errors in prompting and retrieval approaches for ICD-10 coding. We observed that that even when the codes are incorrect, they are often very close. Additionally, MedCodER can overcome prompting and retrieval shortcomings due to its re-ranking capability. 

\section{UI Design and Deployment}

Unlike fully automated ICD coding solutions, MedCodER is an AI-assisted coding tool to enhance medical coding workflows. To illustrate this, we designed a preliminary but functional UI, which is illustrated in Fig.~\ref{fig:main}. For each predicted diagnosis, a button in the UI is available to highlight the corresponding text spans containing disease mentions and supporting evidence texts (captured through the \emph{Extraction} component of our framework). Additionally, a dropdown menu displays MedCodER's top five most relevant ICD-10 codes per diagnosis. Coders can review the highlighted texts and select a code from the dropdown or input a different code or a list of codes as comma-separated values. The application is currently hosted on internal servers and is in the beta-phase. We plan to train coders on usage of the tool and further stress test the application before deploying it to the production environment. For future research, we aim to use the second batch of our annotated dataset (comprising 20 records) along with real-world medical records to conduct A/B testing of the coding application with and without the AI assistant, assessing gains in efficacy and efficiency.

\section{Discussion}
The primary objective of MedCodER is to leverage SOTA LLMs and generative AI techniques for automated medical coding. We focused on GPT-4 in our experiments, however, our framework is not limited to GPT-4 and can work with LLMs such as such as Anthropic’s Claude models or Mistral AI’s Mistral models. Future work will explore other LLMs, including those specific to the biomedical domain, as MedCodER relies on the LLM's knowledge of diseases, supporting evidence, and ICD-10 codes. We anticipate that LLMs trained in the biomedical domain will enhance performance. 

MedCodER assumes medical records are text, but they are often PDFs requiring conversion, which can be challenging with handwritten sections, tables, and other structured data. Additionally, the fixed context length of LLMs may necessitate extra pre-processing steps for longer records. 

Privacy is crucial when using closed-source LLM APIs as medical records contain confidential information. Users must review the terms and conditions of such LLMs. The use of LLMs in healthcare is rapidly evolving, requiring attention to scientific and legal developments. 

\section{Conclusions}

In this paper, we introduce MedCodER, a framework that surpasses existing SOTA medical coding frameworks while maintaining interpretability and a dataset to aid the development of interpretable ICD coding methodologies. We analyzed each component individually and demonstrated that SOTA performance is achieved by combining all components. Additionally, our error analysis highlighted areas for further improvement. Finally, we show how the MedCodER framework can be integrated into an AI-based assistant for medical coders, enhancing their efficiency and accuracy. 

\bibliography{aaai25}

\end{document}